\documentclass{article}
\usepackage{ijcai20}

\usepackage{times}
\usepackage{soul}
\usepackage{url}
\usepackage[hidelinks]{hyperref}
\usepackage[utf8]{inputenc}
\usepackage[small]{caption}
\usepackage{graphicx}
\usepackage{amsmath}
\usepackage{amsthm}
\usepackage{booktabs}
\usepackage{algorithm}
\usepackage{algorithmicx}
\usepackage[noend]{algpseudocode}

\usepackage{amsmath}
\usepackage{amssymb}
\usepackage{subfig}

\title{PENet: Object Detection using Points Estimation in Aerial Images}

\author{
Ziyang Tang$^1$
\and
Xiang Liu$^1$\and
Guangyu Shen$^1$ \And
Baijian Yang$^1$
\affiliations
Purdue University 
\emails
\{tang385, xiang35, shen447, byang\}@purdue.edu
}

\begin{document}

\maketitle
\begin{abstract}
    Aerial imagery has been increasingly adopted in mission-critical tasks, such as traffic surveillance, smart cities, and disaster assistance. However, identifying objects from aerial images faces the following challenges: 1) objects of interests are often too small and too dense relative to the images; 2) objects of interests are often in different relative sizes; and 3) the number of objects in each category is imbalanced. A novel network structure, \textit{Points Estimated Network (PENet)}, is proposed in this work to answer these challenges. \textit{PENet} uses a \textit{Mask Resampling Module (MRM)} to augment the imbalanced datasets, a coarse anchor-free detector (\textit{CPEN}) to effectively predict the center points of the small object clusters, and a fine anchor-free detector \textit{FPEN} to locate the precise positions of the small objects. An adaptive merge algorithm \textit{Non-maximum Merge (NMM)} is implemented in \textit{CPEN} to address the issue of detecting dense small objects, and a hierarchical loss is defined in \textit{FPEN} to further improve the classification accuracy. Our extensive experiments on aerial datasets visDrone \cite{zhuvisdrone2018} and UAVDT \cite{du2018unmanned} showed that PENet achieved higher precision results than existing state-of-the-art approaches. Our best model achieved $8.7\%$ improvement on visDrone and $20.3\%$ on UAVDT.
\end{abstract}

\section{Introduction}
\begin{figure}[tb]
  \centering
  \subfloat[large and sparse objects]{\includegraphics[width=0.45\linewidth]{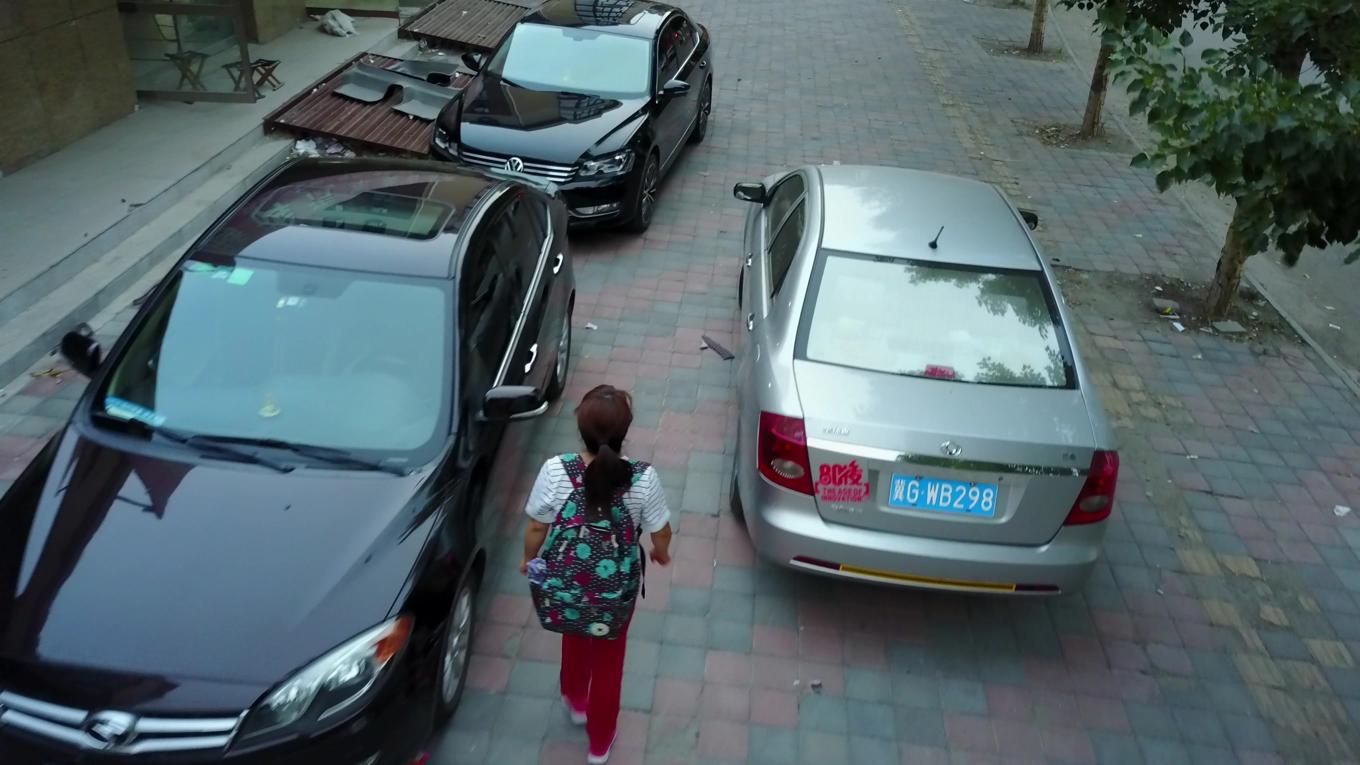}
  \label{fig:largecars}
    }
    \hfill
    \subfloat[small and dense objects]{\includegraphics[width=0.45\linewidth]{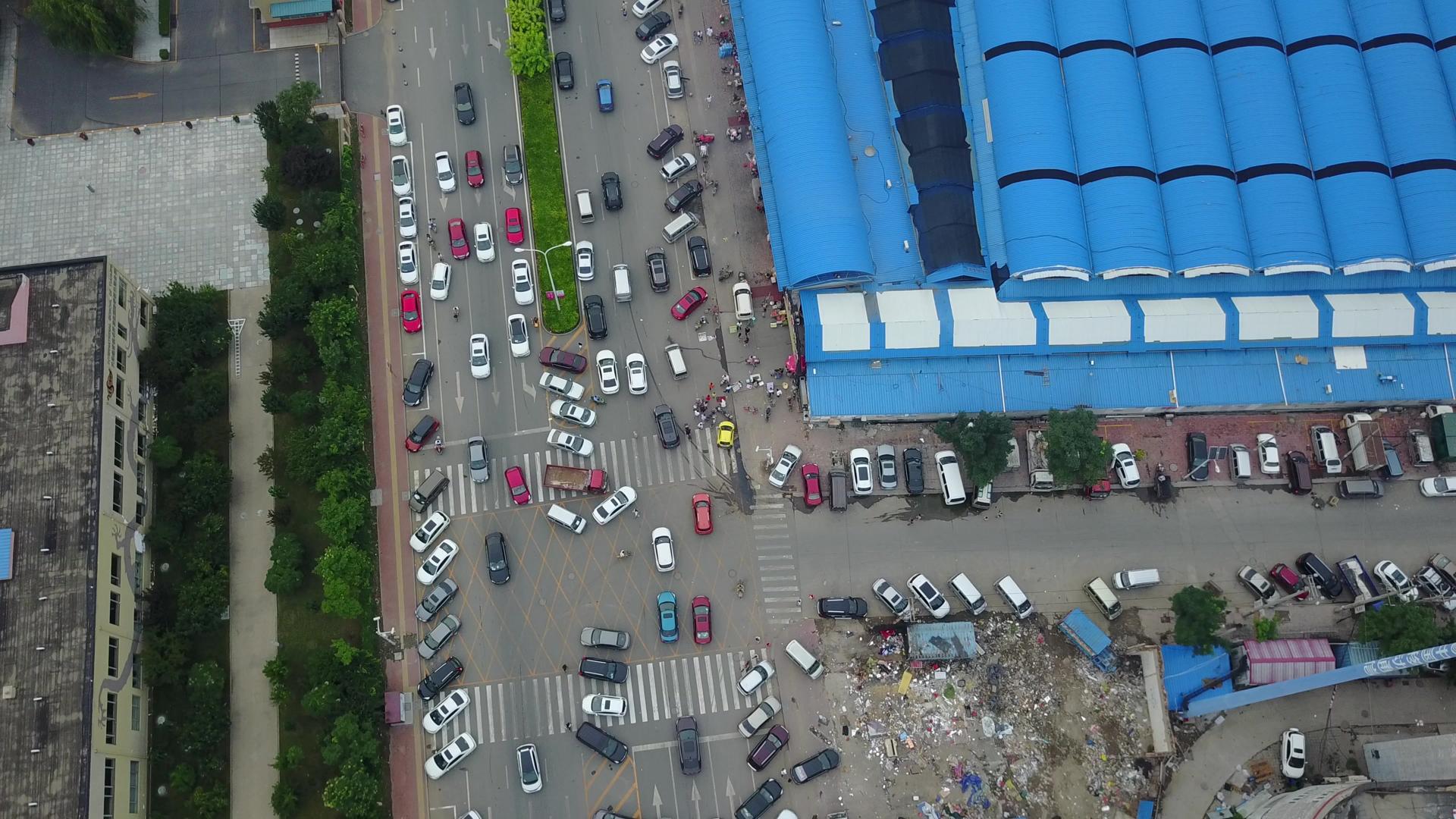}
    \label{fig:smallcars}
    }
  \caption{
  The size of the objects varies from small to large due to different elevations. In (b), when recording from a high perspective, the objects are small and dense in the high resolution image, making it difficult to detect all the objects in aerial images.}
  \label{fig:demo}
\end{figure}

Unmanned Aerial Vehicles (UAV) equipped with a high-resolution camera can be applied to a wide range of applications, including traffic surveillance, smart cities assistance, and disaster respond and recovery. These aerial images are different from the natural imagery datasets, such as COCO \cite{lin2014microsoft}, ImageNet \cite{deng2009imagenet} and Pascal VOC \cite{everingham2010pascal} in the following perspectives:
(1) the aerial images are in high-resolution, e.g. the images in visDrone \cite{zhuvisdrone2018} are about $2,000 \times 1,500$ pixels, while most of the images are less than $500 \times 500$ pixels in COCO \cite{lin2014microsoft}; 
(2) the aerial dataset contains both small, dense objects and sparse, large objects. This is because the elevation of the UAVs changes when collecting the images. 
As shown in Figure \ref{fig:demo}, the size of cars in Figure \ref{fig:smallcars} and Figure \ref{fig:largecars} varies from extremely small to relatively large;
(3) current public aerial imagery datasets are fairly imbalanced with certain types of objects poorly represented.   
For instance, the total number of \textit{cars} in visDrone \cite{zhuvisdrone2018} are \textit{144,866}, while the number of \textit{awning tricycle} in the same datasets is merely \textit{3,246}. In addition, different categories are not equally distinctive: some share more similarities than others. \textit{e.g.} In visDrone \cite{zhuvisdrone2018} dataset, the  \textit{pedestrian} category shares many features with the  \textit{people} category but has little in common with the  \textit{cars} category. In the rest of this paper, we denote this phenomenon as \textit{Categorical Similarity Problems (CSP)}.
\begin{figure*}[tb]
    \centering
    \includegraphics[width=\linewidth]{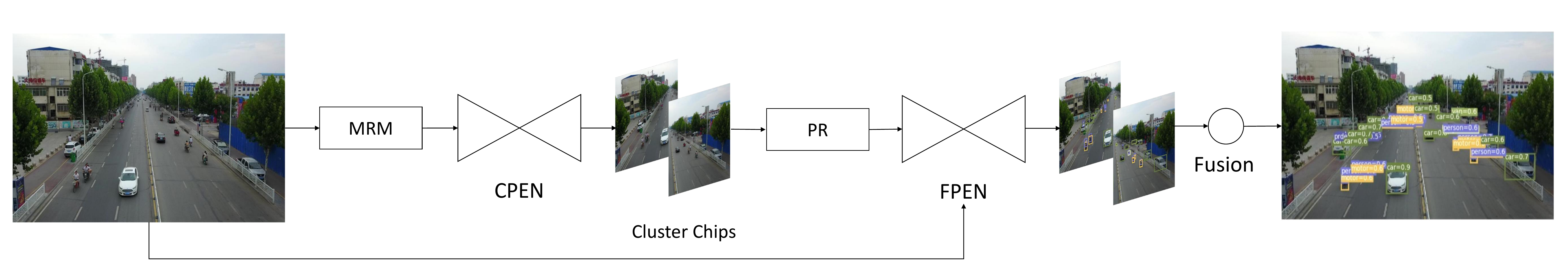}
    \caption{The Overall structure of \textit{PENet}.
    \textit{MRM} augments small objects in the images while 
    \textit{CPEN}  locates the cluster chips from the original high-resolution images. A position refinement is then applied to the candidate cluster chips and the output is fed into \textit{FPEN} which also takes the input of the original image. The prediction results of \textit{FPEN} are fused together as the final detection results.}
    \label{fig:structure}
\end{figure*}

Theses differences between aerial images and natural images makes it  challenging to detect objects in aerial images. Recently, CNN-based detectors demonstrated better performance than the handcrafted detectors \cite{moranduzzo2014detecting}. However, these CNN detectors, including R-CNN families \cite{girshick2015fast,he2017mask}, Yolo and SSD families \cite{redmon2016you,pedoeem2018yolo,liu2016ssd} 
and CornerNet families \cite{law2018cornernet,zhou2019objects}, are all trained from natural images. Due to the aforementioned differences, directly applying the CNN detectors to aerial images will not achieve promising results. In particular, the lack of good processing strategy in high-resolution images and insufficient samples of certain categories are the main reasons to be accounted for.   

"Coarse-to-fine" pipeline was proposed to deal with the problem of resolution and size difference. When the object size is large, the detectors can detect it even after the high-resolution images are down-sampled. When the object sizes are small, the detectors can first find the sub-regions, denoted as \textit{clusters}, that contains a series of small objects, and detect the small objects based on the \textit{clusters}.
The pipeline is a trade-off of performance and computation efficiency. \cite{unelpower} split the images evenly to show the power of tiling in small object detection. \cite{gao2018dynamic} proposed a dynamic zoom-in strategy to speed up the detection with Reinforcement Learning. \cite{yang2019clustered} used K-means to generate high-quality annotations to train a network to find better clusters. To relieve the symptoms of insufficient data samples, \cite{chen2019rrnet,zhang2019fully} reused the samples from the original images to augment the dataset. 

In this paper, we propose the novel \textit{PENet} structure to detect objects in aerial images. Illustrated in Figure \ref{fig:structure}, PENet has three components: \textit{Mask Re-sampling Module (MRM)}, \textit{Coarse-PENet (CPEN), and Fine-PENet (FPEN)}. 
\textit{MRM} is added between the raw images and \textit{CPEN} to augment object samples, especially those small confusing objects. 
\textit{CPEN} generates the center points of high-quality clusters of small objects using an algorithm \textit{Non Maximun Merge (NMM)}. The cluster chips are then fed into \textit{FPEN} for further precise detection. Meanwhile, the original images are also forwarded to \textit{FPEN} to detect large objects from the aerial images. 
The output from \textit{FPEN} are combined to provide the final detection results. When training \textit{FPEN}, a hierarchical loss is applied to address the \textit{CSP}. In addition, the hierarchical loss makes it possible to jointly train a robust model for different aerial datasets.

Our contributions are as follows:
\begin{itemize}
    \item We proposed a novel \textit{PENet} to detect objects of various sizes from high resolution and imbalanced datasets.
    \item We presented  an 
    re-sampling algorithm \textit{MRM}, a cluster generating algorithm \textit{NMM} and a hierarchical loss approach.  Each of them can be independently applied to CNN-based detectors.
    \item We achieved state-of-the-art results in visDrone \cite{zhuvisdrone2018} and UAVDT \cite{du2018unmanned}.
\end{itemize}

\section{Related Works}
In this section, we review data augmentation methods, benchmarks of object detectors, both anchor-based and anchor-free, followed by recent explorations in aerial images. Due to page limit, we focus on the most relevant work and highlight the differences between the proposed solution and the existing approaches.
\subsection{Data Augmentation}
To enlarge dataset, researchers have implemented many methods to augment the data, such as random crop, flipping, and inputs with multi-scale. In aerial imagery datasets, an observation is that the instances from different categories are imbalanced. To mitigate the issues caused by the imbalanced data, \cite{kisantal2019augmentation} pasted small objects to random positions on the training images to improve performance. \cite{chen2019rrnet} improved this approach by considering perspective and reasonableness. They argued that the pasted car instance should be on the road instead of "flying" in the sky. Therefore they proposed \textit{Adasampling} to logistically re-sample the instances. They also discovered that segmentation results are noisy if pre-trained models from natural image datasets are used unmodified. Eroding and median filters were used in their work to partially remove the noises. In contrast, we annotated the segmentation labels and fine-tuned the segmentation models from the aerial images datasets. The high-quality segmentation results act as masks and provide prior information to enable one-step small object pasting. As a result, our \textit{MRM} can quickly generate augmented data with better quality.

\subsection{CNN-based Object Detectors}
The CNN-based object detectors can be divided into two major types: anchor-based detectors and anchor-free detectors. The anchor-based detectors use generated anchor candidates to locate the objects and use a fixed IoU to suppress the negative anchors. In recent years, the most representative anchor-based object detectors are the Faster-RCNN families \cite{girshick2015fast,he2017mask}, 
YoLo families \cite{redmon2016you,redmon2017yolo9000}, 
and SSD families \cite{liu2016ssd}. 
However, the anchor-based detectors rely on good prior anchor sizes, which are often 'inherited' from different datasets. Problem arises when the ratio of the height and the width is significantly different, making the prior anchor size hard to be determined. In addition, most anchor candidates are negative, leading to an imbalance between negative and positive anchors and resulting poor performance in final detection. To address the imbalance in negative and positive anchor candidates, RetinaNet \cite{lin2017focal} introduced \textit{focal loss} by revising the standard cross-entropy loss.

The anchor-free detectors detect the objects based on point estimation. DeNet \cite{tychsen2017denet} estimates the object locations by the top-left, bottom-left or bottom-right corner of a bounding box. It then generates the corner combinations and classifies the anchors using two-stage methods. Point Linking Network (PIL) \cite{wang2017point} predicts the four locations as well as the center of the objects. CornerNet \cite{law2018cornernet} predicts the top-left and bottom-right corners of the bounding boxes and uses corner pooling to combine the corners to get the results. 
Keypoint estimation \cite{zhou2019objects} predicts the corners or the center points and achieved better performance than previous methods. 

Our work adopted anchor-free detectors as the backbone. This is because it is difficult to find good prior anchor sizes that work for all the objects when they have different relative sizes.  \cite{lalonde2018clusternet} also showed that it is easier to locate extremely small objects with points instead of anchors. Our further experiments also show that CNN detectors based on point estimation achieve better performance in aerial datasets.

\subsection{Object Detection in Aerial Image}
The challenge of detecting objects in aerial images is to locate the small objects in high-resolution images. Ultimately, the goal is to make it fast and accurate. However, to the best of our knowledge, none of the existing detectors can achieve this goal. A more realistic approach is to find a better trade-off between accuracy and computational efficiency. The workflow is to find low-resolution image chips from the original images; apply aforementioned detectors; and combine the results. Therefore, the task is to minimize the number of clusters while maintaining good accuracy. \cite{gao2018dynamic} used a reinforcement learning strategy to adaptively zoom in the sub-regions for fine detection. \cite{yang2019clustered} designed a clusDet network and partitioned a high-resolution image into multiple small image chips for detection.  \cite{unelpower} implemented the power of tiling for small object detection.
While the existing methods are predominantly anchor-based, our proposed anchor-free \textit{PENet} is different from all of them. Unlike  \cite{gao2018dynamic}, our \textit{CPEN} can adapatively generate clusters for different images in the dataset. Compared to \cite{yang2019clustered}, our design does not need ROI pooling and scalenet. Instead, our \textit{CPEN} can adapatively generate the clusters and remove some overlapped clusters with position refinement. \cite{unelpower} splits an image evenly with fixed sizes, while our approach offers a few options for the users to choose as a trade-off between performance and computation efficiency: the fastest version generates fewer image chips, while the best model achieves better accuracy.

\section{Methodology}
\subsection{Overview}
\begin{figure}[tb]
    \centering
    \includegraphics[width=\linewidth]{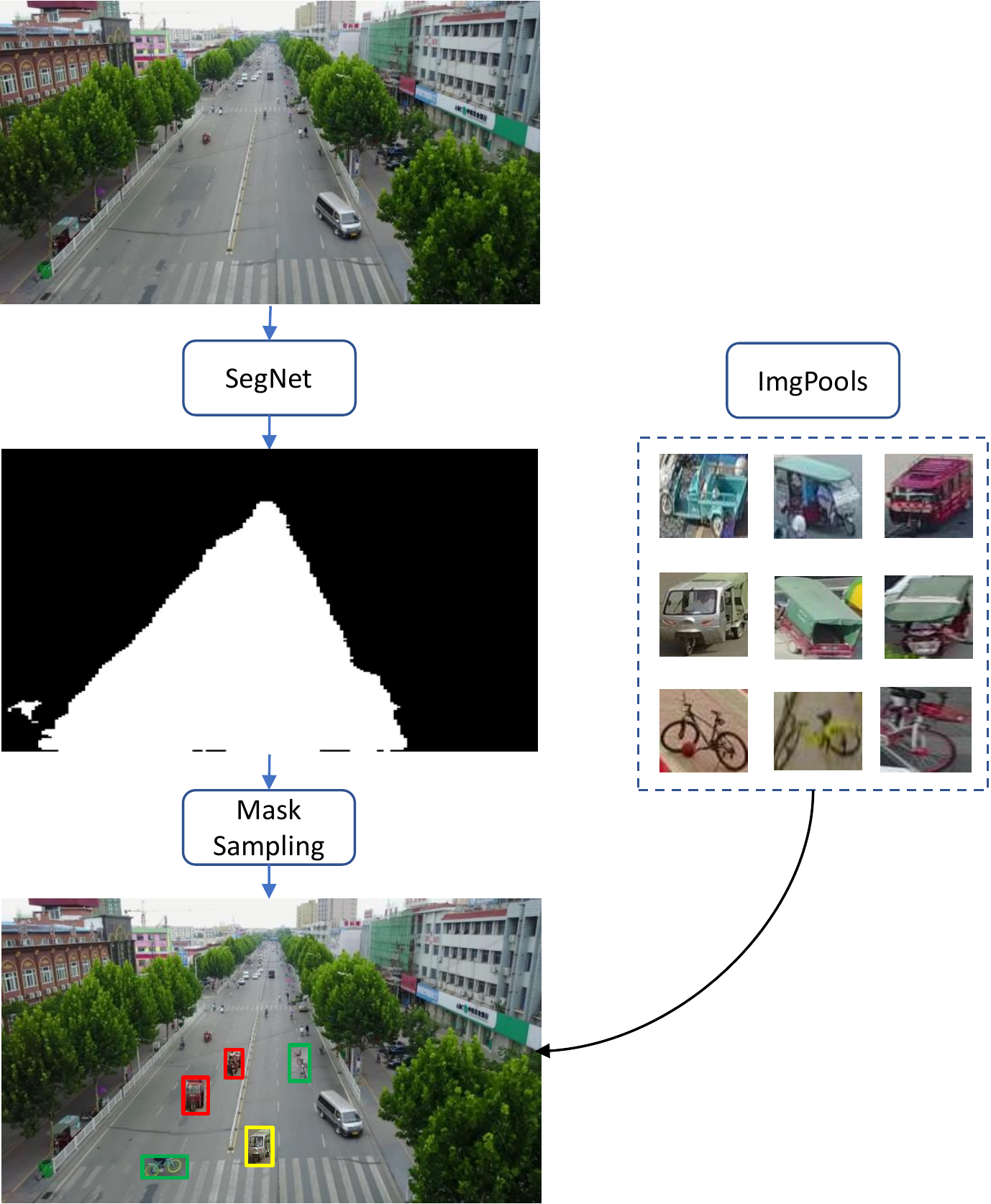}
    \caption{Pipeline of \textit{MRM}. The segmentation network is a fine-tuned network on aerial image datasets, and image pools contain all the confusing objects from the categories in the training data. The \textit{MRM} will find the proper position to paste the objects into the original images.}
    \label{fig:segnet}
\end{figure}
The structure of \textit{PENet} is shown in Figure \ref{fig:structure}. We first use a \textit{Mask Resampling Module (MRM)} for data augmentation. We then apply \textit{Non-Maximum Merge (NMM)} algorithm to merge the small objects into several clusters of similar sizes. Through \textit{Coarse Points Estimation Net (CPEN)}, we get the candidate center points and the size of the clusters. The cluster chips are then processed by \textit{position refinement (PR)} and results are fed into \textit{Fine Points Estimation Net (FPEN)} to detect the center points and the sizes of small objects. The \textit{FPEN} also employs hierarchical loss
for better classification results. Large objects can be detected by using standard down-sampling approach by feeding the image direct into \textit{FPEN}. The final results are obtained by fusing the local results from all the cluster chips and the global results from the original image.

\subsection{MRM}
One of the problems in current aerial images datasets is the imbalanced categories. For example, VisDrone \cite{zhuvisdrone2018} has \textit{144,866} \textit{cars} instances in total, while the number of \textit{awning tricycle} is merely \textit{3,246}. Without appropriate pre-processing, our experiments demonstrated that the accuracy on \textit{cars} are far better than the accuracy on \textit{awning tricycles}. Previous work in RRnet  \cite{chen2019rrnet} showed that randomly re-sample the images may lead to background mismatch and scale mismatch. They proposed to use a pre-trained segmentation network to generate proper positions for data augmentation. However, the network is trained on natural datasets and the discrepancy between aerial images and natural images making the augmented data noisy and undesirable. In our work, we annotated the segmentation labels for the aerial images, and fine-tuned the segmentation networks for aerial images, resulting in far better road maps.In the experiment, we annotated about \textit{500} images from the aerial datasets, and use the fine-tuned model to predict the segmentation results. In addition, we also provided add-on masks of the ground truth to determine proper positions for the augmented objects. \textit{MRM} generates a pool of confusing object images from the whole dataset. Objects are randomly selected from this image pool and pasted to each input image. Since the results of this segmentation network serve as a mask, we named this augmentation component \textit{Mask Re-sampling Module (MRM)}. As Figure \ref{fig:segnet} shows, the \textit{MRM} pipeline uses the segmentation network as a mask. It then re-samples the confusing images from the image pool to synthesize the augmented data. In the experiments, we collected the confusing images from the training data and paste five objects onto each image.

\subsection{CPEN}
\begin{algorithm}[tb]
\caption{Non Max Merge}
\label{alg:NNM}
\textbf{Input}: Sorted bounding boxes $\mathcal{B}=\{B_i\}_{i=1}^{N_\mathcal{B}}$,
classes of the bounding boxes, desired bounding box height $h_b$ and width $w_b$, non max merge threshold $\tau$ \\
\textbf{Output}: Merged clusters bounding boxes $\mathcal{C}$
\begin{algorithmic}[1] 
    \State $\mathcal{C} = \{\}$
    \For{$i \gets 1$ to $N_\mathcal{B}$}
        \If{$B_i$ is \textbf{visited}}
            \State \textbf{continue}
        \EndIf
        \State Flag $B_i$ as \textbf{visited}
        \State $C_i \gets \text{RECENTER}(B_i, h_b, w_b)$
        \For{$j = i+1$ to $N_\mathcal{B}$}
            \If{$\text{IoU}(C_i, B_j) > \tau$}
                \State $\mathcal{C} \gets \mathcal{C} \cup \{B_j\}$
                \State Flag $B_j$ as \textbf{visited}
            \EndIf
        \EndFor
    \EndFor
    \Return $\mathcal{C}$
\end{algorithmic}
\end{algorithm}

\begin{figure}[tb]
    \centering
    \includegraphics[width=\linewidth]{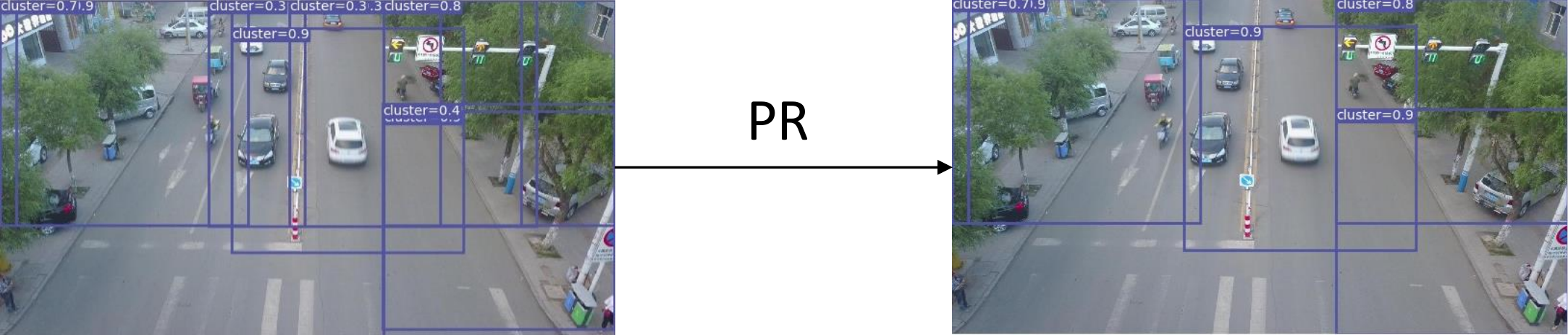}
    \caption{An example of the position refinement to get rid of some candidate clusters. In this example, we can get rid of 5 candidate clusters from the top 10 prediction results.}
    \label{fig:cluster}
\end{figure}

The goal of \textit{CPEN} is to train a model that can estimate high-quality clusters of small objects. In previous work, \cite{yang2019clustered} proposed a k-means algorithm to generate the ground truth of the clusters with each image having the same number of the clusters. However, in aerial images, the number of the clusters can be non-identical because of the difference in camera perspective, shooting time and other situations. Therefore a fixed number of clusters is not a good representative of all the situations. In this paper, we proposed \textit{Non-maximum Merge (NMM)} algorithm to adaptively generate different number of clusters for different images. Shown in Algorithm \ref{alg:NNM}, starting from the top-left boxes, we apply a cluster $\mathcal{C}_i$ with size of $(w,h)$ and merge all the small objects $\mathcal{B}$ in $\mathcal{C}_i$. This process is repeated until all the objects are merged. $\mathcal{C}$ is the set with all the clusters $\mathcal{C}_i$. Note that some objects may be cropped into two parts during the merge process. Therefore, we set up a threshold $\tau$ to allow a slight overlap among the clusters. It could happen that the clusters may go beyond the boundary of the original images. Under this circumstance, the RECENTER function relocates the center of $C_i$ to ensure the whole cluster is still in the image. 
During the inference stage, we predict the $topk$ center points of the clusters. Some clusters may share a large overlap with others. These candidate clusters will be removed by the \textit{position refinement (PR)} operation. Figure \ref{fig:cluster} shows that the number of the clusters is reduced from 10 to 5 after \textit{PR}. It is evident that \textit{PR} helps to speed up the inference time while maintaining good performance. The trade-off of performance and computation time can be made by carefully choosing the value of $topk$ and the cluster overlap threshold $\tau$. 

\subsection{FPEN}
\textit{FPEN} is inspired by the hierarchical classification from YOLO9000 \cite{redmon2017yolo9000}, where a word tree strategy is used to handle classes with \textit{CSP} issues. In the aerial dataset like visDrone \cite{zhuvisdrone2018}, the classes have similar \textit{CSP} problems. For example, \textit{people} class and \textit{pedestrian} class share more similarities than \textit{people} class and \textit{cars} class. Therefore, we propose a hierarchical loss in our anchor-free detectors. The backbone network can be any detectors in the CornerNet series  \cite{law2018cornernet,zhou2019objects,Man+18}. In our experiments, we implemented our hierarchical loss based on CenterNet \cite{zhou2019objects}:
\begin{align}
L_{shm} = \frac{-1}{N}\sum
\begin{cases}
(1-\hat{Y})^\alpha log(\hat{Y}) & \text{if } Y=1\\
(1-Y)^\beta (\hat{Y})^\alpha log(1-\hat{Y}) & \text{otherwise}
\end{cases}
\end{align}
Where $Y\in [0,1]^{\frac{W}{R}\times \frac{H}{R} \times (C+C_s)} $. $W,H$, and $C$ represents width, height and number of classes, respectively. $R$ is the ratio for down-sampling. $C_s$ is the stacked classes. Followed the setting in  \cite{zhou2019objects}, we set $\alpha=2$, $\beta=4$ and $R=4$. The settings of $C$ and $C_s$ are different for each dataset. For example, in visDrone  \cite{zhuvisdrone2018}, $C=11$ and $C_s=3$ with additional classes of \textit{human}, \textit{vehicles} and \textit{non-motor-vehicles}. To generate the bounding box from the predicted center points, we also implement a regression model with the $L_{wh}$ and $L_{off}$:
\begin{align}
    L_{wh} = \frac{1}{N}\sum_{k=1}^{N} |\hat{s}_{k} - s_k|\\
    L_{off} = \frac{1}{N}\sum_{p}|\hat{O}_p - (\frac{p}{R} - \lfloor \frac{p}{R} \rfloor)|
\end{align}
, where $N$ is the total number of the centerpoints, and $s$ represents the size of the bounding boxes. Therefore, the object function can be summarized as follows:
\begin{align}
    L = \lambda_{shm} L_{shm} + \lambda_{wh} L_{wh} + \lambda_{off} L_{off}
\end{align}
In our experiments, we set $\lambda_{shm} =1, \lambda_{wh}=0.1, \lambda_{off}=1$.
\subsection{Results Fusion}
Final outputs are generated by merging the detection results from each cluster chip. 
To predict large objects, we feed-forward original images using standard down-sampling operations. The merge process is a standard Non Maximum Suppression. In some cases, an object may be cropped into two parts in different clusters and therefore have multiple bounding boxes. This can be fixed by merging the two bounding boxes into one larger box.

\section{Experiments}
\begin{figure*}[tb]
    \centering
    \subfloat{\includegraphics[width=0.3\linewidth]{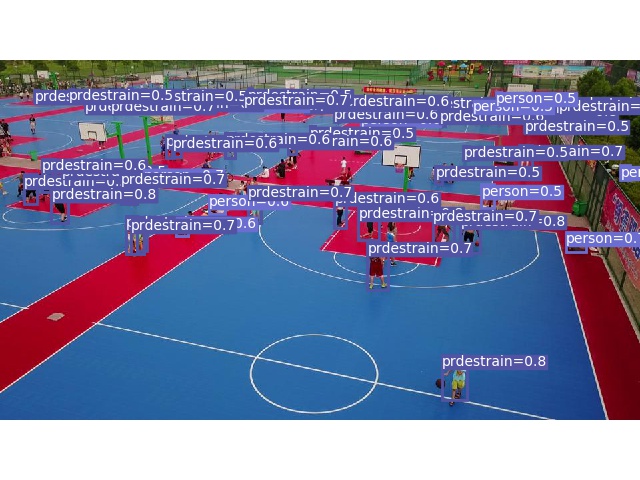}
    }
    \hfill
    \subfloat{\includegraphics[width=0.3\linewidth]{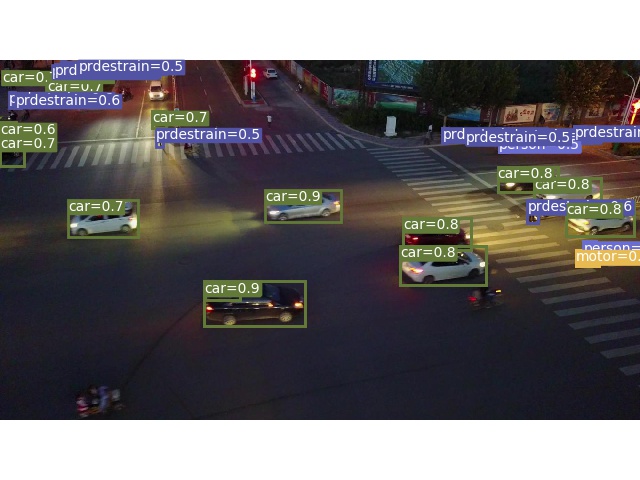}
    }
    \hfill
    \subfloat{\includegraphics[width=0.3\linewidth]{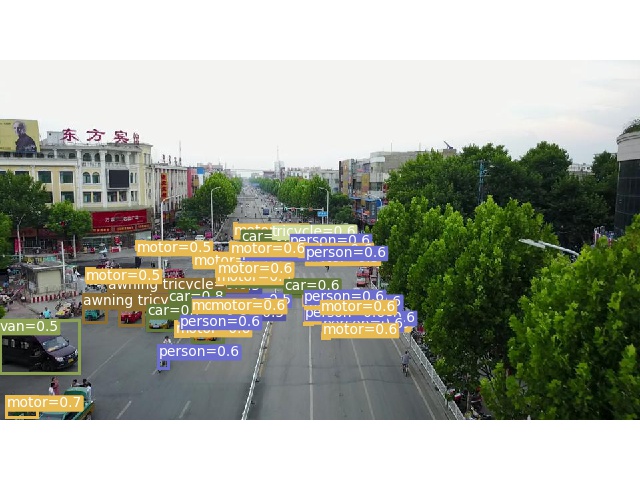}
    }
    \hfill
    \subfloat{\includegraphics[width=0.3\linewidth]{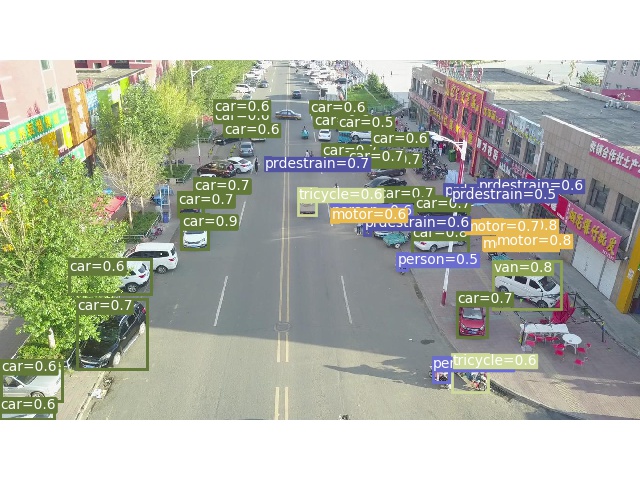}
    }
    \hfill
    \subfloat{\includegraphics[width=0.3\linewidth]{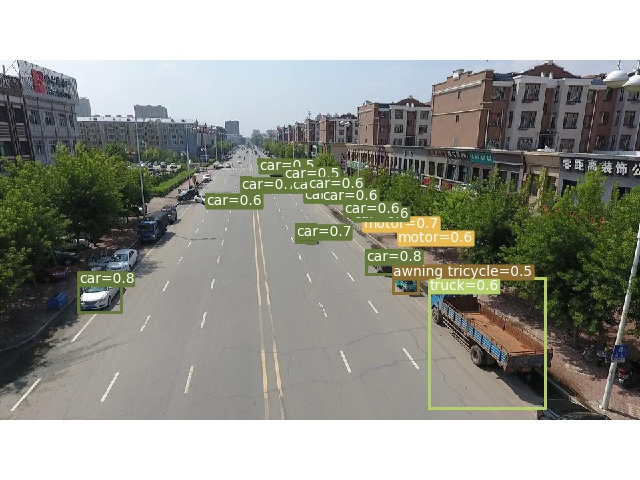}
    }
    \hfill
    \subfloat{\includegraphics[width=0.3\linewidth]{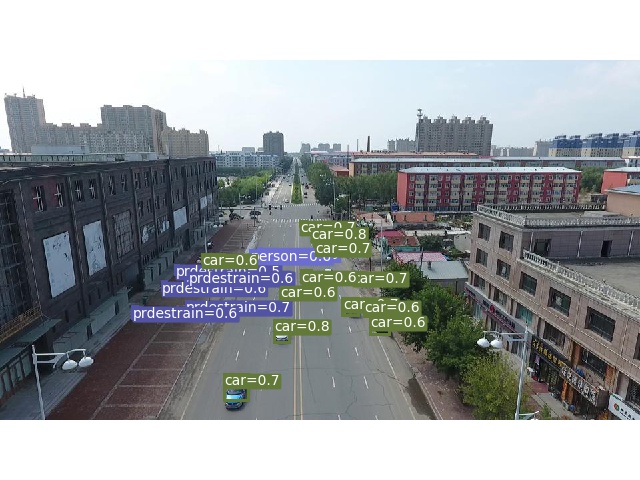}
    }
    \hfill
    \caption{Some visualization results from our \textit{PENet}. The performance is good at finding both the small and large objects.}
    \label{fig:results}
\end{figure*}
We implemented the proposed \textit{PENet} on Pytorch 1.13. The models are trained and tested with two RTX 2080Ti GPUs. In the training phase, we trained our models for 200 epochs, with the learning rate starting from $1.25e^{-4}$, decaying by 10 separately at epoch 90 and 120. Two public aerial image datasets: visDrone \cite{zhuvisdrone2018} and UAVDT \cite{du2018unmanned} are used in our study. In visDrone, we chose $Top10$ cluster centers and chop the image chips into \textit{FPEN}. We then chose $Top100$ object center points for each image chip as the detection results. Whereas in UAVDT, we chose $top5$ in \textit{CPEN} and $top100$ in \textit{FPEN}. The maximum detected objects is set to 500. The $\tau$ is set to $0.8$ for all our experiments.
We configured the same settings from \cite{zhou2019objects} and used maxpooling instead of standard NMS for evaluation. Figure \ref{fig:results} shows a few examples of the outputs of \textit{PENet}.

\subsection{Evaluate Metrics}
We adopted standard  $AP$ criterion from  \cite{lin2014microsoft} to evaluate the performance of our methods against other benchmarks. AP is the average of all \textit{10} $IoU$ thresholds range from $[0.5:0.95]$ with step size of $0.05$ of all classes. $AP50$ and $AP75$ are computed with the IOU threshold set to \textit{0.5}, and \textit{0.75} of all categories, respectively.  
\subsection{VisDrone}
\begin{table}[tb]
\centering
\caption{The comparison of the performance in visDrone}
\label{table:visdrone}
\footnotesize {  
\resizebox{.5\textwidth}{!}{
\begin{tabular}{c|ccc|cccc}
\toprule
Methods      & AP{[}\%{]} & AP50{[}\%{]} & AP75{[}\%{]} \\ \midrule
RetineaNet \cite{lin2017focal}   & 11.81      & 21.37        & 11.62\\
RefineDet \cite{zhang2018single}    & 14.90      & 28.76        & 14.08 \\
DetNet59 \cite{li2018detnet}     & 15.26      & 29.23        & 14.34 \\
FPN \cite{lin2017feature}          & 16.51      & 32.20        & 14.91 \\
Light-RCNN \cite{li2017light}   & 16.53      & 32.78        & 15.13 \\
CornetNet \cite{law2018cornernet}    & 17.41      & 34.12        & 15.78 \\
Cascade-RCNN \cite{cai2018cascade} & 16.09      & 31.91        & 15.01 \\\cline{1-4}
RRNet \cite{chen2019rrnet}       & 29.62     &54.00      &28.70\\
ClusDet \cite{yang2019clustered}                         & 32.4      & 56.2          & 31.6\\ \cline{1-4}
CenterNet \cite{zhou2019objects}   &14.2 &19.3 &15.5 \\
CPEN+CenterNet \cite{zhou2019objects}                            &29.3  &38.7   &32.4\\
CPEN+FPEN &36.7 &53.2 &39.5\\
CPEN+FPEN+MRM &\textbf{41.1} &\textbf{58.0} &\textbf{44.3} \\ \bottomrule
\end{tabular}
}
} 
\end{table}

The datasets consists \textit{6,471} training images and \textit{548} validation images. 
Unfortunately, the server is shut down and we were not able to evaluate our results on the test set. Instead, validation set are used to evaluate different models. The whole dataset contains \textit{12} classes: \textit{ignored regions, pedestrian, person, bicycle, car, van, truck, tricycle, awning tricycle, bus, motor, others}. To implement hierarchical loss, we added three additional classes: \textit{Human, non-vehicles}, and \textit{vehicles}. The average resolution of the images is about $2,000 \times 1,500$.  Table \ref{table:visdrone} shows \textit{CPEN} improves $15\%$ than the baseline;  \textit{FPEN} has a $7\%$ improvement; and \textit{MRM} is seen $5\%$ increasement. Our best model increased $8.7\%$ in mAP than the state-of-the-art detectors. It is worth noting that our proposed \textit{PENet} has better performance in $AP75$, indicating that our predictions are closer to the ground truth. A possible explanation for better results achieved by our model is that our prediction is based on the center points of the object and thus the predicted bounding boxes have bigger overlap with the ground truth.

\subsection{UAVDT}
UAVDT is a video aerial dataset that contains three categories of objects: \textit{car, bus}, and \textit{truck}. It has \textit{80,000} annotated frames from \textit{100} videos. The video sequences cover different weather conditions, including daylights, night and fog. The dataset also records from different perspectives, such as the perspective of front, side or bird view. The resolution of the frame is $1,024 \times 540$ on average. The comparison of the performance is shown in table \ref{tab:uavDT}. Comparing against existing work, our best proposed \textit{PENet} improves the performance by $20.6\%$ than the state-of-the-art. One explanation is that the previous work are anchor-based detectors. In such a challenging dataset with different weather conditions and distinct shooting views, it is really difficult to find a limited number of prior anchors to represent all the object with different sizes. The glamouring results proved that our point estimation approach is a good fit for detecting small objects in high resolution images.
\begin{table}[tb]
\centering
\caption{The comparison of the performance in UAVDT}
\label{tab:uavDT}
\resizebox{.5\textwidth}{!}{
\begin{tabular}{c|ccc}
\toprule
Methods      & AP{[}\%{]} & AP50{[}\%{]} & AP75{[}\%{]} \\ \midrule
SSD \cite{liu2016ssd}    & 9.3      & 21.4        & 6.7 \\
Faster-RCNN \cite{girshick2015fast} + FPN \cite{lin2017feature}     &11.0      &23.4        &8.4 \\
ClusDet \cite{yang2019clustered}          &13.7      &26.5        &12.5 \\
DREN \cite{zhang2019fully}         &17.7 &N/A &N/A \\ \cline{1-4}
LSN \cite{wang2019learning} &31.9 &51.4 &33.6 \\
NDFT \cite{Wu_2019_ICCV} &46.62 &N/A &N/A \\
\cline{1-4}
CenterNet \cite{zhou2019objects} &23.5  &41.0   &24.5\\
CPEN + CenterNet \cite{zhou2019objects} &62.6 &69.0 &67.8\\
CPEN+MRM &\textbf{67.3} &\textbf{76.3} &\textbf{74.6} \\ \bottomrule
\end{tabular}
}
\end{table}


\subsection{Ablation Study}
In this section, we perform an ablation study on the performance of each component of  \textit{PENet}.
\begin{table*}[tb]
\centering
\caption{Albation Study on VisDrone.}
\label{tab:ablation}
\resizebox{1\textwidth}{!}{
\begin{tabular}{c|cccccccccc}
\toprule
Methods & \textit{prdestrain} & \textit{person} & \textit{bicycle} & \textit{car} & \textit{van} & \textit{truck} & \textit{tricycle} & \textit{awning tricycle} & \textit{bus} & \textit{motor} \\ \midrule
BaseLine &10.97 &8.11 &1.94 &46.50 &18.77 &14.97 &5.81 &2.35 &22.55 &10.09 \\
CPEN & 37.43 & 22.24 & 18.96 & 60.78 & 31.59 & 22.29 & 16.88 & 8.55 & 45.99 & 27.82\\
CPEN + FPEN & 43.46 & 28.03 & 22.35 & 67.42 & 43.47 & 33.41 & 28.23 & 13.09 & 49.34 & 38.29\\ 
CPEN + FPEN + MRM &\textbf{47.84} &\textbf{34.53} &\textbf{26.45} &\textbf{72.29} &\textbf{45.30} &\textbf{37.23} &\textbf{31.37} &\textbf{17.14} &\textbf{54.16} &\textbf{44.35}  \\ \bottomrule
\end{tabular}
}
\end{table*}
\subsubsection{CPEN}
\textit{CPEN} uses the proposed \textit{NMM} algorithm to adaptively generate the center points of the clusters. By providing more precise cluster chips, \textit{CPEN} enjoys a significant improvement over the baseline of CenterNet \cite{zhou2019objects}. In visDrone, \textit{CPEN} improved by $15.1\%$ compared to the baseline. In UAVDT, \textit{CPEN} improved $38.1\%$. From Table \ref{tab:ablation}, the mAP of class \textit{bicycle} increased significantly from \textit{1.94} to \textit{18.96}. The great improvement indicates the importance of extracting the clusters chips when detecting  small object in aerial images. Note that the improvement comes from the high quality cluster chips generated from our \textit{CPEN}, and our \textit{CPEN} can be extended as an additional component for all other types of CNN-based object detectors, including anchor-free detectors, e.g. CenterNet  \cite{zhou2019objects}, as well as anchor-based detectors, e.g. Faster-RCNN \cite{girshick2015fast}, YOLO \cite{redmon2016you} or SSD \cite{liu2016ssd}.

\subsubsection{FPEN}
\textit{FPEN} employs hierarchical loss to address the \textit{CSP}. Shown in Table \ref{table:visdrone}, it is seen a $7.4\%$ improvement in AP. The improvements of mAP in each category are also evaluated. Since UAVDT data has only three categories, It is not very beneficial to apply hierarchical loss. Results of visDrone dataset, shown in Table \ref{tab:ablation}, reveals that adding hierarchical loss can improve the mAP for all the categories, especially those small objects, e.g. \textit{bicycle} and \textit{tricycle}. The promising results demonstrated the importance of addressing the \textit{CSP} in CNN-based detectors. Another noteworthy advantage is that hierarchical loss makes it possible to train a model with multiple different datasets. As a data-driven technique, more data from different distribution will empower CNN-based detectors to be more robust and to have better generalization ability.
\subsubsection{MRM}
As aforementioned, we extract the road maps and logistically add instance patches. In the experiments, we added five instances in each input image. Table \ref{tab:ablation} shows \textit{MRM} improved the mAP by up to $4.4\%$ in the final performance of visDrone data and $4.7\%$ in UAVDT data. The results show the value of augmenting data in improving the final performance. Note that \textit{MRM} is a generalized data augmentation approach and can be implemented on other CNN-based object detectors as an alternative of re-sampling modules.

\section{Conclusion}
In this paper, we presented a novel object detector \textit{PENet} for object dections in aerial image datasets. \textit{CPEN} can estimate the most relevant cluster chips. \textit{FPEN} can predict the precise location of small objects and address the importance of \textit{CSP} in detection tasks. \textit{MRM} provides an alternative re-sampling method as a data augmentation approach. Our experiments showed that \textit{PENet} achieved state-of-the-art performance in two aerial imagery datasets, visDrone \cite{zhuvisdrone2018} and UAVDT \cite{du2018unmanned}.

\bibliographystyle{named}
\small{ \bibliography{ijcai20} }

\begin{thebibliography}{}

\bibitem[\protect\citeauthoryear{Cai and Vasconcelos}{2018}]{cai2018cascade}
Zhaowei Cai and Nuno Vasconcelos.
\newblock Cascade r-cnn: Delving into high quality object detection.
\newblock In {\em Proceedings of the IEEE conference on computer vision and
  pattern recognition}, pages 6154--6162, 2018.

\bibitem[\protect\citeauthoryear{Chen \bgroup \em et al.\egroup
  }{2019}]{chen2019rrnet}
Changrui Chen, Yu~Zhang, Qingxuan Lv, Shuo Wei, Xiaorui Wang, Xin Sun, and
  Junyu Dong.
\newblock Rrnet: A hybrid detector for object detection in drone-captured
  images.
\newblock In {\em Proceedings of the IEEE International Conference on Computer
  Vision Workshops}, pages 0--0, 2019.

\bibitem[\protect\citeauthoryear{Deng \bgroup \em et al.\egroup
  }{2009}]{deng2009imagenet}
Jia Deng, Wei Dong, Richard Socher, Li-Jia Li, Kai Li, and Li~Fei-Fei.
\newblock Imagenet: A large-scale hierarchical image database.
\newblock In {\em 2009 IEEE conference on computer vision and pattern
  recognition}, pages 248--255. Ieee, 2009.

\bibitem[\protect\citeauthoryear{Du \bgroup \em et al.\egroup
  }{2018}]{du2018unmanned}
Dawei Du, Yuankai Qi, Hongyang Yu, Yifan Yang, Kaiwen Duan, Guorong Li, Weigang
  Zhang, Qingming Huang, and Qi~Tian.
\newblock The unmanned aerial vehicle benchmark: Object detection and tracking.
\newblock In {\em Proceedings of the European Conference on Computer Vision
  (ECCV)}, pages 370--386, 2018.

\bibitem[\protect\citeauthoryear{Everingham \bgroup \em et al.\egroup
  }{2010}]{everingham2010pascal}
Mark Everingham, Luc Van~Gool, Christopher~KI Williams, John Winn, and Andrew
  Zisserman.
\newblock The pascal visual object classes (voc) challenge.
\newblock {\em International journal of computer vision}, 88(2):303--338, 2010.

\bibitem[\protect\citeauthoryear{Gao \bgroup \em et al.\egroup
  }{2018}]{gao2018dynamic}
Mingfei Gao, Ruichi Yu, Ang Li, Vlad~I Morariu, and Larry~S Davis.
\newblock Dynamic zoom-in network for fast object detection in large images.
\newblock In {\em Proceedings of the IEEE Conference on Computer Vision and
  Pattern Recognition}, pages 6926--6935, 2018.

\bibitem[\protect\citeauthoryear{Girshick}{2015}]{girshick2015fast}
Ross Girshick.
\newblock Fast r-cnn.
\newblock In {\em Proceedings of the IEEE international conference on computer
  vision}, pages 1440--1448, 2015.

\bibitem[\protect\citeauthoryear{He \bgroup \em et al.\egroup
  }{2017}]{he2017mask}
Kaiming He, Georgia Gkioxari, Piotr Doll{\'a}r, and Ross Girshick.
\newblock Mask r-cnn.
\newblock In {\em Proceedings of the IEEE international conference on computer
  vision}, pages 2961--2969, 2017.

\bibitem[\protect\citeauthoryear{Kisantal \bgroup \em et al.\egroup
  }{2019}]{kisantal2019augmentation}
Mate Kisantal, Zbigniew Wojna, Jakub Murawski, Jacek Naruniec, and Kyunghyun
  Cho.
\newblock Augmentation for small object detection.
\newblock {\em arXiv preprint arXiv:1902.07296}, 2019.

\bibitem[\protect\citeauthoryear{LaLonde \bgroup \em et al.\egroup
  }{2018}]{lalonde2018clusternet}
Rodney LaLonde, Dong Zhang, and Mubarak Shah.
\newblock Clusternet: Detecting small objects in large scenes by exploiting
  spatio-temporal information.
\newblock In {\em Proceedings of the IEEE Conference on Computer Vision and
  Pattern Recognition}, pages 4003--4012, 2018.

\bibitem[\protect\citeauthoryear{Law and Deng}{2018}]{law2018cornernet}
Hei Law and Jia Deng.
\newblock Cornernet: Detecting objects as paired keypoints.
\newblock In {\em Proceedings of the European Conference on Computer Vision
  (ECCV)}, pages 734--750, 2018.

\bibitem[\protect\citeauthoryear{Li \bgroup \em et al.\egroup
  }{2017}]{li2017light}
Zeming Li, Chao Peng, Gang Yu, Xiangyu Zhang, Yangdong Deng, and Jian Sun.
\newblock Light-head r-cnn: In defense of two-stage object detector.
\newblock {\em arXiv preprint arXiv:1711.07264}, 2017.

\bibitem[\protect\citeauthoryear{Li \bgroup \em et al.\egroup
  }{2018}]{li2018detnet}
Zeming Li, Chao Peng, Gang Yu, Xiangyu Zhang, Yangdong Deng, and Jian Sun.
\newblock Detnet: A backbone network for object detection.
\newblock {\em arXiv preprint arXiv:1804.06215}, 2018.

\bibitem[\protect\citeauthoryear{Lin \bgroup \em et al.\egroup
  }{2014}]{lin2014microsoft}
Tsung-Yi Lin, Michael Maire, Serge Belongie, James Hays, Pietro Perona, Deva
  Ramanan, Piotr Doll{\'a}r, and C~Lawrence Zitnick.
\newblock Microsoft coco: Common objects in context.
\newblock In {\em European conference on computer vision}, pages 740--755.
  Springer, 2014.

\bibitem[\protect\citeauthoryear{Lin \bgroup \em et al.\egroup
  }{2017a}]{lin2017feature}
Tsung-Yi Lin, Piotr Doll{\'a}r, Ross Girshick, Kaiming He, Bharath Hariharan,
  and Serge Belongie.
\newblock Feature pyramid networks for object detection.
\newblock In {\em Proceedings of the IEEE conference on computer vision and
  pattern recognition}, pages 2117--2125, 2017.

\bibitem[\protect\citeauthoryear{Lin \bgroup \em et al.\egroup
  }{2017b}]{lin2017focal}
Tsung-Yi Lin, Priya Goyal, Ross Girshick, Kaiming He, and Piotr Doll{\'a}r.
\newblock Focal loss for dense object detection.
\newblock In {\em Proceedings of the IEEE international conference on computer
  vision}, pages 2980--2988, 2017.

\bibitem[\protect\citeauthoryear{Liu \bgroup \em et al.\egroup
  }{2016}]{liu2016ssd}
Wei Liu, Dragomir Anguelov, Dumitru Erhan, Christian Szegedy, Scott Reed,
  Cheng-Yang Fu, and Alexander~C Berg.
\newblock Ssd: Single shot multibox detector.
\newblock In {\em European conference on computer vision}, pages 21--37.
  Springer, 2016.

\bibitem[\protect\citeauthoryear{Maninis \bgroup \em et al.\egroup
  }{2018}]{Man+18}
K.K. Maninis, S.~Caelles, J.~Pont-Tuset, and L.~{Van Gool}.
\newblock Deep extreme cut: From extreme points to object segmentation.
\newblock In {\em Computer Vision and Pattern Recognition (CVPR)}, 2018.

\bibitem[\protect\citeauthoryear{Moranduzzo and
  Melgani}{2014}]{moranduzzo2014detecting}
Thomas Moranduzzo and Farid Melgani.
\newblock Detecting cars in uav images with a catalog-based approach.
\newblock {\em IEEE Transactions on Geoscience and Remote Sensing},
  52(10):6356--6367, 2014.

\bibitem[\protect\citeauthoryear{Pedoeem and Huang}{2018}]{pedoeem2018yolo}
Jonathan Pedoeem and Rachel Huang.
\newblock Yolo-lite: a real-time object detection algorithm optimized for
  non-gpu computers.
\newblock {\em arXiv preprint arXiv:1811.05588}, 2018.

\bibitem[\protect\citeauthoryear{Redmon and Farhadi}{2017}]{redmon2017yolo9000}
Joseph Redmon and Ali Farhadi.
\newblock Yolo9000: better, faster, stronger.
\newblock In {\em Proceedings of the IEEE conference on computer vision and
  pattern recognition}, pages 7263--7271, 2017.

\bibitem[\protect\citeauthoryear{Redmon \bgroup \em et al.\egroup
  }{2016}]{redmon2016you}
Joseph Redmon, Santosh Divvala, Ross Girshick, and Ali Farhadi.
\newblock You only look once: Unified, real-time object detection.
\newblock In {\em Proceedings of the IEEE conference on computer vision and
  pattern recognition}, pages 779--788, 2016.

\bibitem[\protect\citeauthoryear{Tychsen-Smith and
  Petersson}{2017}]{tychsen2017denet}
Lachlan Tychsen-Smith and Lars Petersson.
\newblock Denet: Scalable real-time object detection with directed sparse
  sampling.
\newblock In {\em Proceedings of the IEEE International Conference on Computer
  Vision}, pages 428--436, 2017.

\bibitem[\protect\citeauthoryear{Unel \bgroup \em et al.\egroup }{}]{unelpower}
F~Ozge Unel, Burak~O Ozkalayc{\i}, and Cevahir C{\i}gla.
\newblock The power of tiling for small object detection.

\bibitem[\protect\citeauthoryear{Wang \bgroup \em et al.\egroup
  }{2017}]{wang2017point}
Xinggang Wang, Kaibing Chen, Zilong Huang, Cong Yao, and Wenyu Liu.
\newblock Point linking network for object detection.
\newblock {\em arXiv preprint arXiv:1706.03646}, 2017.

\bibitem[\protect\citeauthoryear{Wang \bgroup \em et al.\egroup
  }{2019}]{wang2019learning}
Tiancai Wang, Rao~Muhammad Anwer, Hisham Cholakkal, Fahad~Shahbaz Khan, Yanwei
  Pang, and Ling Shao.
\newblock Learning rich features at high-speed for single-shot object
  detection.
\newblock In {\em Proceedings of the IEEE International Conference on Computer
  Vision}, pages 1971--1980, 2019.

\bibitem[\protect\citeauthoryear{Wu \bgroup \em et al.\egroup
  }{2019}]{Wu_2019_ICCV}
Zhenyu Wu, Karthik Suresh, Priya Narayanan, Hongyu Xu, Heesung Kwon, and
  Zhangyang Wang.
\newblock Delving into robust object detection from unmanned aerial vehicles: A
  deep nuisance disentanglement approach.
\newblock In {\em The IEEE International Conference on Computer Vision (ICCV)},
  October 2019.

\bibitem[\protect\citeauthoryear{Yang \bgroup \em et al.\egroup
  }{2019}]{yang2019clustered}
Fan Yang, Heng Fan, Peng Chu, Erik Blasch, and Haibin Ling.
\newblock Clustered object detection in aerial images.
\newblock {\em arXiv preprint arXiv:1904.08008}, 2019.

\bibitem[\protect\citeauthoryear{Zhang \bgroup \em et al.\egroup
  }{2018}]{zhang2018single}
Shifeng Zhang, Longyin Wen, Xiao Bian, Zhen Lei, and Stan~Z Li.
\newblock Single-shot refinement neural network for object detection.
\newblock In {\em Proceedings of the IEEE Conference on Computer Vision and
  Pattern Recognition}, pages 4203--4212, 2018.

\bibitem[\protect\citeauthoryear{Zhang \bgroup \em et al.\egroup
  }{2019}]{zhang2019fully}
Junyi Zhang, Junying Huang, Xuankun Chen, and Dongyu Zhang.
\newblock How to fully exploit the abilities of aerial image detectors.
\newblock In {\em Proceedings of the IEEE International Conference on Computer
  Vision Workshops}, pages 0--0, 2019.

\bibitem[\protect\citeauthoryear{Zhou \bgroup \em et al.\egroup
  }{2019}]{zhou2019objects}
Xingyi Zhou, Dequan Wang, and Philipp Kr{\"a}henb{\"u}hl.
\newblock Objects as points.
\newblock {\em arXiv preprint arXiv:1904.07850}, 2019.

\bibitem[\protect\citeauthoryear{Zhu \bgroup \em et al.\egroup
  }{2018}]{zhuvisdrone2018}
Pengfei Zhu, Longyin Wen, Xiao Bian, Ling Haibin, and Qinghua Hu.
\newblock Vision meets drones: A challenge.
\newblock {\em arXiv preprint arXiv:1804.07437}, 2018.

\end{thebibliography}

\end{document}